\documentclass[runningheads]{llncs}
\usepackage{graphicx,xcolor}

\usepackage{tabu}
\usepackage{booktabs}       % professional-quality tables
\usepackage{amsfonts}       % blackboard math symbols
\usepackage{nicefrac}       % compact symbols for 1/2, etc.
\usepackage{microtype}      % microtypography
\usepackage{amsmath}
\usepackage{mathtools}
\usepackage{latexsym}
\usepackage{multirow}
\usepackage{xspace}
\usepackage{tabularx}

\usepackage{color}
\usepackage{amsmath, amssymb}
\usepackage{type1cm}
\usepackage{comment}
\usepackage{amsmath}
\usepackage{amsfonts}
\usepackage{bm}
\usepackage{mathtools}
\usepackage{tikz}
\usepackage[left]{lineno}
\newcommand{\argmin}{\mathop{\rm argmin}\limits}

\newcommand{\Rset}{\mathbb{R}}

\newcommand{\transpose}{^{\mathrm{T}}}
\newcommand{\diag}{\mathop{\text{diag}}}

\let\mytablefont\scriptsize

\begin{document}
%\linenumbers
%
\title{Binarized Knowledge Graph Embeddings}
%\iffalse
\author{$^\dagger$ Koki Kishimoto\inst{1}
%\orcidID{0000-1111-2222-3333} 
\and
Katsuhiko Hayashi\inst{1,3}
%\orcidID{1111-2222-3333-4444} 
\and
Genki Akai\inst{1}
%\orcidID{2222--3333-4444-5555}
\and \\
Masashi Shimbo\inst{2,3} 
\and
Kazunori Komatani\inst{1}
%$^\dagger$\url{kishimoto@ei.sanken.osaka-u.ac.jp}
}
\authorrunning{K. Kishimoto et~al.}
\institute{
Osaka University, Osaka, Japan
\and
Nara Institute of Science and Technology, Nara, Japan
\and
RIKEN Center for Advanced Intelligence Project\\
$^\dagger$\url{kishimoto@ei.sanken.osaka-u.ac.jp}
}
%
%\fi

\maketitle              % typeset the header of the contribution
\begin{abstract}
Tensor factorization has become an increasingly 
popular approach to knowledge graph completion~(KGC), which
is the task of automatically predicting missing facts in a knowledge graph.
However,
% since exsiting knowledge graphs are rapidly growing in scale,
even with a simple model
like 
%CP decomposition,
CAN\-DECOMP\slash PARA\-FAC (CP) %~\cite{cp}
tensor decomposition,
KGC on existing knowledge graphs
is impractical in resource-limited environments,
as
a large amount of memory
is required to store % the 32-bit or 64-bit
parameters
represented as 32-bit or 64-bit floating point numbers.
This limitation is expected to become more stringent
as existing knowledge graphs, which are already huge, keep steadily growing in scale.
%% The required number of parameters is proportional
%% to that of vertices in a graph,
%% and many knowledge graphs, which already boast billions of edges, are rapidly growing in scale.
%%
%% To deal with large-scale knowledge graphs,
%% it is desirable to
%% significantly reducing memory footprints of
%% factorization models.
%Quantization is recognized as one of the most
%effective methods to satisfy the extreme memory requirements
%in deep neual network models.
To reduce the memory requirement,
we present a method for
binarizing the parameters of the CP tensor decomposition
by introducing a quantization function
to the optimization problem.
This method replaces floating point--valued parameters
%% used at the time of KCG
with binary ones after training, which drastically reduces
the model size at run time.
%% This method replaces floating point--valued parameters
%% with binary ones, which drastically reduces
%% the size of the model used for KGC.
%%
%This method can drastically
%reduce the model size
%by replacing floating point values
%with binary ones.
We investigate the trade-off between 
the quality and size of tensor factorization models
for several KGC benchmark datasets.
In our experiments,
the proposed method 
successfully reduced the model size
by more than an order of magnitude
while maintaining the task performance.
Moreover, a fast score computation technique can be
developed with bitwise operations.
%% which opens the possibility of performing massively parallel KGC using many resource-limited devices.


%%% LocalWords: KGC CP binarize binarizing

%%% Local Variables:
%%% mode: latex
%%% TeX-PDF-mode: t
%%% TeX-engine: xetex
%%% TeX-master: "main"
%%% End:

\end{abstract}
\section{Introduction}
\label{sec:intro}
%Large-scale
%\memo{Large-scale(Abstractの表現に合わせた)}
%knowledge graphs, such as YAGO~\cite{yago} and Freebase~\cite{freebase},
%are indispensable resource for knowledge-intensive
%applications such as question answering~\cite{qa}, dialog~\cite{diag}
%and recommender~\cite{rec} systems.
Knowledge graphs, such as YAGO~\cite{yago} and Freebase~\cite{freebase},
have proven useful in many applications such as question answering~\cite{qa}, dialog~\cite{diag}
and recommender~\cite{rec} systems.
%Knowledge graphs encode structured information of entities and their rich relations.
%A knowledge graph is a collection of
%triples $(e_i,e_j,r_k)$,
%each of which represents a fact that relation $r_k$ holds between subject entity
%$e_i$ and object entity $e_j$.
A knowledge graph consists of triples $(e_i,e_j,r_k)$,
%,where $e_i$ and $e_j$ 
%The facts are stored of home as triple $(e_i,e_j,r_k)$
each of which represents a fact that relation $r_k$ holds between subject entity
$e_i$ and object entity $e_j$.
Although a typical knowledge graph may have billions of
triples, it is still far from complete.
%If a knowledge graph is complete, in the sense
%that it contains all necessary facts for applications,
%it would be possible to answer various questions
%by carrying out inference
%%\memo{by searching, by carrying out searching
%%知識グラフが完全なら推論というより検索の方が正しい？}
%on the knowledge graph alone.
%However, existing knowledge graphs are incomplete and many facts are missing~\cite{survey}.
Filling in the missing triples is of importance in carrying out
various inference over knowledge graphs.
\emph{Knowledge graph completion}~(KGC) aims to perform this task automatically.
% find such missing facts automatically.
%This is important to predict unknown facts.

In recent years, \emph{knowledge graph embedding} (KGE) 
has been actively pursued as a promising approach to KGC.
In KGE, entities and relations are embedded
in vector space,
and operations in this space are used for
defining a confidence score (or simply score) function $\theta_{i j k}$
that approximates the truth value of a given triple $(e_i,e_j,r_k)$.
Although a variety of KGE methods~\cite{transe,rescal,ntn,complex,conve}
have been proposed,
Kazemi and Poole~%(2018)~
\cite{simple} and
Lacroix et~al.~%(2018)~
\cite{cano} found
that 
a classical tensor factorization algorithm,
known as the 
CAN\-DECOMP\slash PARA\-FAC (CP) 
decomposition~\cite{cp},
achieves the state-of-art performances on several benchmark datasets for KGC.

In CP decomposition of a knowledge graph, the confidence score $\theta_{i j k}$ for
a triple $(e_i,e_j,r_k)$ is calculated simply by 
%$\bm{a}_i\transpose(\bm{b}_j\circ\bm{c}_k)$
$\bm{a}_{i:}(\bm{b}_{j:}\circ\bm{c}_{k:})\transpose$
where $\bm{a}_{i:}$, $\bm{b}_{j:}$, and $\bm{c}_{k:}$ denote the $D$-dimensional
row vectors representing $e_i$, $e_j$, and $r_k$, respectively,
and $\circ$ is the Hadamard~(element-wise) product.
%Entity and relation vectors require significant amounts of memory and storage.
In spite of the model's simplicity, it
needs to maintain $(2N_e+N_r)$ $D$-dimensional 32-bit or 64-bit valued
vectors, where $N_e$, and $N_r$ denote the number
of entities and relations, respectively.
Because typical knowledge graphs contain enormous number of entities and relations,
this leads to
a significant memory requirement.
As mentioned in~\cite{conve},
%% actually, a simple KGE model like
CP with $D=200$ applied to Freebase
will require about 66 GB of memory to store parameters.
This large memory consumption poses issues especially when
KGC is conducted on resource-limited devices.
%\memo{in the case of employing a KGE model, when a KGE model is used
%employing a KGE model}
%a KGE model especially on resource limited devices.
Moreover, the size of existing knowledge graphs is still growing rapidly, and
a method for shrinking the embedding vectors is in strong demand.

To address the problem, this paper presents a new CP decomposition
algorithm to learn compact knowledge graph embeddings.
The basic idea is to introduce
a quantization function built into the optimization problem.
This function forces the embedding vectors to be binary,
and optimization is done with respect to the binarized vectors.
After training, the binarized embeddings can be used in place of the original vectors of floating-point numbers,
which drastically reduces the memory footprint of the resulting model.

In addition, the binary vector representation
contributes to efficiently computing
the dot product % between binary vectors
by using bitwise operations.
%% instead of standard floating point unit operations.
This fast computation % of dot products
allows
the proposed model to substantially
reduce the amount of time
required to compute the confidence scores of triples.

Note that our method only improves the run-time (i.e., predicting missing triples) memory footprint and speed but not those for training a prediction model.
However,
the reduced memory footprint of the produced model 
enables KGC to be run on many affordable resource-limited devices (e.g., personal computers).
Unlike research-level benchmarks in which one is required to compute the scores of a small set of test triples,
completion of an entire knowledge graph requires computing the scores of all missing triples in a knowledge graph,
whose number is enormous because knowledge graphs are sparse.
Thus, improved memory footprints and reduced score computation time are of practical importance,
and these are what our proposed model provides.

The quantization technique has been commonly used in the community
of deep neural networks to shrink network components~\cite{bicon,surveyq}.
To the best of our knowledge, this technique has not been studied
in the field of tensor factorization.
The main contribution of this paper is that
we introduce the quantization function 
to a tensor factorization model for the first time.
This is also the first study to investigate the benefits
of the quantization for KGC. 
Our experimental results on several KGC benchmark datasets
showed that the proposed method
reduced the model size nearly 10- to 20-fold
compared to the standard CP decomposition
without a decrease in the task performance.
Besides, with bitwise operations,
B-CP got a bonus of 
%2-3 times 
speed-up in score computation time.
%\memo{In addition, the quantization method can
%conduct KGC quickly with bitwise operation,
%which enables \emph{high-speed inference} in large-scale knowledge graph.}

%%%%%%%%%%%%%%%%%%%%%%%%%%%%%%%%%%%%%%%%%%%%%%%%%%%%%%%%%%%%%%%%%%%%%%%%

%%% LocalWords: KGC KGE YAGO Kazemi Lacroix et al CANDECOMP PARAFAC CP Hadamard binarized

%%% Local Variables:
%%% mode: latex
%%% TeX-PDF-mode: t
%%% TeX-engine: xetex
%%% TeX-master: "main"
%%% End:

\section{Related Work}
%\subsection{Knowledge graph embeddings}
Approaches to knowledge graph embedding~(KGE)
can be classified into three types: models
based on bilinear mapping, translation, and neural network-based transformation.

RESCAL~\cite{rescal} is a bilinear-based KGE method whose score function
is formulated as $\theta_{ijk}=\bm{a}_{e_i}^{\rm T}\bm{B}_{r_k}\bm{a}_{e_j}$,
where $\bm{a}_{e_i}, \bm{a}_{e_j} \in \Rset^D$ are the vector representations
of entities $e_i$ and $e_j$, respectively,
and matrix $\bm{B}_{r_k} \in \Rset^{D\times D}$ represents a relation $r_k$.
%They make it possible to capture the asymmetry property by the vector representation matrix of relations being an asymmetry matrix.
Although RESCAL is able to output non-symmetric score functions,
each relation matrix $\bm{B}_{r_k}$ holds $D^2$ parameters.
This can be problematic both in terms of overfitting and computational cost.
To avoid this problem, several methods have been proposed recently.
DistMult~\cite{distmult} restricts the relation matrix to be
diagonal, $\bm{B}_{r_k} = \diag(\bm{b}_{r_k})$.
However, this form of function is necessarily symmetric in $i$ and $j$;
i.e., $\theta_{ijk}=\theta_{jik}$.
To reconcile efficiency and expressiveness,
Trouillon et~al.~(2016)~\cite{complex}
proposed ComplEx, %which can also be regarded as a special case of RESCAL.
using the complex-valued representations and
Hermitian inner product to define the score function,
which unlike DistMult, can be nonsymmetric in $i$ and $j$.
%%$\mathop{\text{Re}} \left( \mat{e}_s\transpose  \diag(\mat{w}_r) \overline{\mat{e}_o} \right) $.
Hayashi and Shimbo~(2017)~\cite{eq} found that ComplEx is equivalent to another state-of-the-art KGE method,
holographic embeddings~(HolE)~\cite{hole}.
ANALOGY~\cite{analogy} is a model that can be view as a hybrid of ComplEx and DistMult.
Manabe et~al.~(2018)~\cite{l1} reduced redundant parameters of ComplEx
with L1 regularizers.
Lacroix et~al.~(2018)~\cite{cano} and
Kazemi and Pool~(2018)~\cite{simple} independently showed
that CP decomposition~(called SimplE in~\cite{simple}) achieves a comparable
KGC performance to other bilinear methods such as ComplEx and ANALOGY.
To achieve this performance,
they introduced an ``inverse'' triple $(e_j,e_i,r_k^{-1})$ to the training data
for each existing triple $(e_i, e_j, r_k)$,
where $r_k^{-1}$ denotes the inverse relation of $r_k$.

TransE~\cite{transe} is the first KGE model based on vector translation.
It employs the principle ${\bm a}_{e_i}+{\bm b}_{r_k}\approx{\bm a}_{e_j}$
to define a distance-based score function $ \theta_{ijk} = - \|{\bm a}_{e_i}+{\bm b}_{r_k}-{\bm a}_{e_j}\|^{2}$.
Since TransE was recognized as too limited to
model complex properties (e.g., symmetric/reflexive/one-to-many/many-to-one relations) in knowledge graphs,
many extended versions of TransE have been proposed.
%However, it has been recently pointed out that
%some of these models, such as TransH~\cite{transh},
%TransR~\cite{transr}, and STransE~\cite{stranse},
%are still not sufficiently expressive~\cite{simple}.

Neural-based models, such as NTN~\cite{ntn} and ConvE~\cite{conve},
employ non-linear functions to define score function,
and thus they have a better expressiveness.
Compared to bilinear and translation approaches, however,
neural-based models require more complex operations to compute
interactions between a relation and two entities in vector space.

It should be noted that the binarization technique proposed in this paper can be applied
to other KGE models besides CP decomposition, such as those mentioned above.
Our choice of CP as the implementation platform only reflects the fact that
it is one of the strongest baseline KGE methods.

Numerous recent publications have studied methods for training quantized
neural networks to compact the models
without performance degradation.
The BinaryConnect algorithm~\cite{bicon} is the first study to show
that binarized neural networks can achieve almost the state-of-the-art
results on datasets such as MNIST and CIFAR-10~\cite{surveyq}.
BinaryConnect uses the binarization function $Q_1(x)$
to replace floating point weights of deep neural networks
with binary weights during the forward and backward propagation.
Lam~(2018)~\cite{w2b} used the same quantization method as BinaryConnect
to learn compact word embeddings.
To binarize knowledge graph embeddings,
this paper also applied the quantization method to
the CP decomposition algorithm.
To the best of our knowledge, this paper is the first study
to examine the benefits of the quantization for KGC.

%%%%%%%%%%%%%%%%%%%%%%%%%%%%%%%%%%%%%%%%%%%%%%%%%%%%%%%%%%%%%%%%%%%%%%%%

%%% LocalWords: KGC KGE CANDECOMP PARAFAC CP
%%% LocalWords: ComplEx HolE SimplE NTN TransE TransH STransE TransR ConvE
%%% LocalWords: bilinear overfitting nonsymmetric binarize binarized binarization quantize quantized
%%% LocalWords: Hayashi Shimbo Hermitian Kazemi Lacroix et al
%%% LocalWords: MNIST CIFAR BinaryConnect

%%% Local Variables:
%%% mode: latex
%%% TeX-PDF-mode: t
%%% TeX-engine: xetex
%%% TeX-master: "main"
%%% End:

\section{Notation and Preliminaries}
We follow the notation and terminology established in~\cite{tens} for the most part.
These are summarized below mainly for third-order tensors,
by which a knowledge graph is represented
(see Section~\ref{sec:kgrep}).

Vectors are represented by boldface lowercase letters, e.g., $\bm{a}$.
Matrices are represented by boldface capital letters, e.g., $\bm{A}$.
Third-order tensors are represented by boldface calligraphic
letters, e.g., $\bm{\mathcal{X}}$.

The $i$th row of a matrix $\bm{A}$ is represented by $\bm{a}_{i:}$,
and
the $j$th column of $\bm{A}$ is represented by 
$\bm{a}_{:j}$,
%% Alternatively, the $j$th column of a matrix, $\bm{a}_{:j}$, may be represented
or simply as $\bm{a}_j$.
The symbol $\circ$ represents the Hadamard product
for matrices and also for vectors,
and $\otimes$ represents the % vector
outer product.

A third-order tensor $\bm{\mathcal{X}} \in \mathbb{R}^{I_1 \times I_2 \times I_3}$
is rank one if it can be written as the outer product of three vectors, i.e.,
% \begin{displaymath}
$
\bm{\mathcal{X}}=\bm{a} \otimes \bm{b} \otimes \bm{c}
$.
% \end{displaymath}
%% where the symbol $\otimes$ represents the vector
%% outer product.
This means that each element
$ x_{i_1 i_2 i_3} $
of $\bm{\mathcal{X}}$ is the product of the corresponding
vector elements:
\begin{displaymath}
x_{i_1 i_2 i_3}=a_{i_1}b_{i_2}
c_{i_3} \quad \text{for }i_1 \in [I_1]\text{, } i_2 \in [I_2]\text{, }
i_3 \in [I_3] \text{,}
\end{displaymath}
where $[I_n]$ denotes the set of natural numbers $1,2,\cdots,I_n$.

%% The norm of a tensor $\bm{\mathcal{X}} \in \mathbb{R}^{I_1 \times
%% I_2 \times I_3}$
%% is the square root of the sum of the squares
%% of all its elements, i.e.,
%% \begin{displaymath}
%% \|\bm{\mathcal{X}}\|=\sqrt{\sum_{i_1=1}^{I_1}\sum_{i_2=1}^{I_2}\sum_{i_3=1}^{I_3}x_{i_1 i_2i_3}^2}.
%% \end{displaymath}
The norm of a tensor $\bm{\mathcal{X}} \in \mathbb{R}^{I_1 \times
I_2 \times \cdots \times I_k}$
is the square root of the sum of the squares
of all its elements, i.e.,
\begin{displaymath}
%% \|\bm{\mathcal{X}}\|=\sqrt{\sum_{i_1=1}^{I_1}\sum_{i_2=1}^{I_2} \cdots \sum_{i_k=1}^{I_k}x_{i_1 i_2 \cdots i_k}^2}.
\|\bm{\mathcal{X}}\|=\sqrt{\sum_{ i_1 \in [I_1] } \sum_{ i_2 \in [I_2] } \cdots \sum_{ i_k \in [I_k] }x_{i_1 i_2 \cdots i_k}^2}.
\end{displaymath}
For a matrix (or a second-order tensor), this norm
is called the Frobenius norm and is represented by $\|\cdot\|_F$.
%% The $l_1$-norm $\|\bm{A}\|_1$ of a matrix $\bm{A}\in\Rset^{N\times M}$ is given by
%% $\sum_{ i \in [N] }\sum_{ j \in [M] }|a_{ij}|$.
%% %% $\sum_{i=1}^{N}\sum_{j=1}^{M}|a_{ij}|$.

%%% Local Variables:
%%% mode: latex
%%% TeX-PDF-mode: t
%%% TeX-engine: xetex
%%% TeX-master: "main"
%%% End:

\section{Tensor Factorization for Knowledge Graphs}
\subsection{Knowledge Graph Representation}
\label{sec:kgrep}
A knowledge graph $\mathcal{G}$ 
is a labeled multigraph 
$(\mathcal{E},\mathcal{R},\mathcal{F})$,
where $\mathcal{E} = \{e_1,\ldots,e_{N_e}\}$ is the set of entities (vertices),
$\mathcal{R} = \{r_1, \ldots ,r_{N_r}\}$
is the set of all relation types (edge labels), and
$\mathcal{F} \subset \mathcal{E} \times \mathcal{E} \times \mathcal{R}$
denotes the observed instances of relations over entities (edges).
The presence of an edge, or a triple, $(e_i,e_j,r_k) \in \mathcal{F}$ represents the fact 
that relation $r_k$ holds between subject entity $e_i$
and object entity $e_j$. 

A knowledge graph can be represented as a 
boolean third order tensor $\bm{\mathcal{X}}
\in \{0,1\}^{N_e \times N_e \times N_r }$
whose elements are set such as
\begin{displaymath}
x_{ijk} = 
\begin{cases} 
   1 & \text{if } (e_i,e_j,r_k) \in \mathcal{F} \\
   0 & \text{otherwise}
\end{cases}.
\end{displaymath}
KGC is concerned 
with incomplete knowledge graphs, i.e.,
$\mathcal{F} \subsetneq \mathcal{F}^*$,
where $\mathcal{F}^* \subset \mathcal{E} \times \mathcal{E} \times \mathcal{R}$
is the unobservable set of ground truth facts (and a superset of $\mathcal{F}$).
KGE 
has been recognized as a promising approach to predicting
the truth value of unknown triples in $\mathcal{F^*} \setminus \mathcal{F}$.
KGE can be generally formulated as the tensor factorization problem
and defines a score function $\theta_{ijk}$ using
the latent vectors of entities and relations.
%In recent studies~\cite{cano,simple},
%it has been shown that
%among several tensor factorization models,
%the canonical decomposition~(also called 
%CANDECOMP/PARAFAC or CP) is an effective
%tensor factorization model for KGC, despite its simplicity.

%% In the following, we describe the formal framework of 
%% the CP decomposition model for KGC.

\subsection{CP Decomposition}
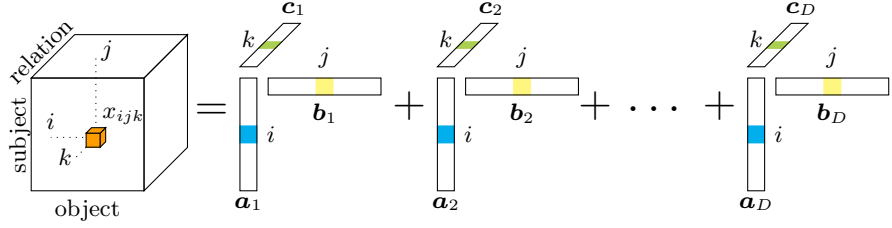
\begin{figure}[t]
\centering
%\begin{figure}[t]
%\centering
\begin{tikzpicture}[scale=0.75]

%\definecolor{lightblue}{HTML}{FF9D00}
%\definecolor{blue}{HTML}{0000ff}
%\definecolor{green}{HTML}{008000}
%\definecolor{red}{HTML}{ff0000}

\definecolor{blue}{HTML}{00AFEC}
%\definecolor{blue}{HTML}{FF9D00}
%orange
\definecolor{red}{HTML}{AACF52}
%green
\definecolor{green}{HTML}{FFF67F}
%yellow
%\definecolor{lightblue}{HTML}{00AFEC}
\definecolor{lightblue}{HTML}{FF9D00}
%lightblue

%\definecolor{lightblue}{HTML}{FF9D00}
%\definecolor{blue}{HTML}{0068B7}
%\definecolor{green}{HTML}{009944}
%\definecolor{red}{HTML}{E60012}
\pgfmathsetmacro{\cubex}{2}
\pgfmathsetmacro{\cubey}{2}
\pgfmathsetmacro{\cubez}{2}
\draw[black] (0,0,0) -- ++(-\cubex,0,0) -- ++(0,-\cubey,0) -- ++(\cubex,0,0) -- cycle;
\draw[black] (0,0,0) -- ++(0,0,-\cubez) -- ++(0,-\cubey,0) -- ++(0,0,\cubez) -- cycle;
\draw[black] (0,0,0) -- ++(-\cubex,0,0) -- ++(0,0,-\cubez) -- ++(\cubex,0,0) -- cycle;

%\node (X) at (-1,-1.2,0) [above] {$\mathcal{X}$} ;

\node (subject) at (-1,-2,0) [below] {object} ;
\node (object) at (-2.2,-1,0) [rotate=90] {subject} ;
\node (relation) at (-1.85,0.5,0) [rotate=45] {relation} ;

\pgfmathsetmacro{\x}{0}
\pgfmathsetmacro{\y}{-0.2}
\pgfmathsetmacro{\z}{0.3}
\pgfmathsetmacro{\cubex}{1.2 - \x}
\pgfmathsetmacro{\cubexx}{0.8}
\pgfmathsetmacro{\cubey}{1.2 - \y}
\pgfmathsetmacro{\cubez}{1.2 - \z}
\draw[black,dotted] (-1.2,-1.2+\y,-1.2+\z) -- ++(-\cubexx,0,0) ;
\draw[black,dotted] (-1.2+\x,0,-1.2+\z) -- ++(0,-\cubey,0) ;
\draw[black,dotted] (-1.2+\x,-1.2+\y,0) -- ++(0,0,-\cubez) ;
\pgfmathsetmacro{\cubex}{0.25}
\pgfmathsetmacro{\cubey}{0.25}
\pgfmathsetmacro{\cubez}{0.25}
\filldraw[draw=black,fill=lightblue] (-1.075+\x,-1.075+\y,-1.075+\z) -- ++(-\cubex,0,0) -- ++(0,-\cubey,0) -- ++(\cubex,0,0) -- cycle;
\filldraw[draw=black,fill=lightblue] (-1.075+\x,-1.075+\y,-1.075+\z) -- ++(0,0,-\cubez) -- ++(0,-\cubey,0) -- ++(0,0,\cubez) -- cycle;
\filldraw[draw=black,fill=lightblue] (-1.075+\x,-1.075+\y,-1.075+\z) -- ++(-\cubex,0,0) -- ++(0,0,-\cubez) -- ++(\cubex,0,0) -- cycle;

\node (relation) at (-1.4+\x,-1.5+\y,0) [above] {$k$} ;
\node (relation) at (-2,-1.2+\y,-1.2+\z) [above] {$i$} ;
\node (relation) at (-1.0+\x,-0.2,-1.2+\z) [above] {$j$} ;

\node (xijk) at (-0.775,-1.375,-1.075) [above] {$x_{ijk}$} ;

\node (equal) at (1.2,-0.9,0) [above,font=\LARGE] {$=$} ;

\pgfmathsetmacro{\cubex}{0.3}
\pgfmathsetmacro{\cubey}{0.3}
\pgfmathsetmacro{\cubez}{2}
\filldraw [draw=blue, fill=blue] (2,-0.85) -- ++(-\cubex,0) -- ++(0,-\cubey) -- ++(\cubex,0) -- cycle;
\pgfmathsetmacro{\cubey}{2}
\draw[black] (2,0) -- ++(-\cubex,0) -- ++(0,-\cubey) -- ++(\cubex,0) -- cycle;

\node (A) at (1.85,-2.5,0) [above] {$\bm{a}_1$} ;
\node (B) at (3.2,-0.3,0) [below] {$\bm{b}_1$} ;
\node (C) at (2.65,1.5,0) [below] {$\bm{c}_1$} ;
\node (i) at (2.25,-0.7,0) [below] {$i$} ;
\node (j) at (3.2,0.7,0) [below] {$j$} ;
\node (k) at (1.85,1.0,0) [below] {$k$} ;

\pgfmathsetmacro{\cubex}{0.3}
\pgfmathsetmacro{\cubey}{0.3}
\pgfmathsetmacro{\cubez}{2}
\filldraw[draw=green, fill=green] (3.35,0) -- ++(-\cubex,0) -- ++(0,-\cubey) -- ++(\cubex,0) -- cycle;

\pgfmathsetmacro{\cubex}{2}
\draw[black] (4.2,0) -- ++(-\cubex,0) -- ++(0,-\cubey) -- ++(\cubex,0) -- cycle;

\pgfmathsetmacro{\cubex}{0.3}
\pgfmathsetmacro{\cubey}{2}
\pgfmathsetmacro{\cubez}{0.3}
\filldraw[draw=red, fill=red] (2,0.2,-0.85) -- ++(-\cubex,0,0) -- ++(0,0,-\cubez) -- ++(\cubex,0,0) -- cycle;
\pgfmathsetmacro{\cubez}{2}
\draw[black] (2,0.2,0) -- ++(-\cubex,0,0) -- ++(0,0,-\cubez) -- ++(\cubex,0,0) -- cycle;

\pgfmathsetmacro{\span}{3.5}
\pgfmathsetmacro{\cubex}{0.3}
\pgfmathsetmacro{\cubey}{0.3}
\pgfmathsetmacro{\cubez}{2}
\filldraw [draw=blue, fill=blue] (\span+2,-0.85) -- ++(-\cubex,0) -- ++(0,-\cubey) -- ++(\cubex,0) -- cycle;
\pgfmathsetmacro{\cubey}{2}
\draw[black] (\span + 2,0) -- ++(-\cubex,0) -- ++(0,-\cubey) -- ++(\cubex,0) -- cycle;

\node (A) at (\span+1.85,-2.5,0) [above] {$\bm{a}_2$} ;
\node (B) at (\span+3.2,-0.3,0) [below] {$\bm{b}_2$} ;
\node (C) at (\span+2.65,1.5,0) [below] {$\bm{c}_2$} ;
\node (i) at (\span+2.25,-0.7,0) [below] {$i$} ;
\node (j) at (\span+3.2,0.7,0) [below] {$j$} ;
\node (k) at (\span+1.85,1.0,0) [below] {$k$} ;

\pgfmathsetmacro{\cubey}{0.3}
\pgfmathsetmacro{\cubez}{2}
\pgfmathsetmacro{\cubex}{0.3}
\filldraw[draw=green, fill=green] (\span+3.35,0) -- ++(-\cubex,0) -- ++(0,-\cubey) -- ++(\cubex,0) -- cycle;
\pgfmathsetmacro{\cubex}{2}
\draw[black] (\span+4.2,0) -- ++(-\cubex,0) -- ++(0,-\cubey) -- ++(\cubex,0) -- cycle;

\pgfmathsetmacro{\cubex}{0.3}
\pgfmathsetmacro{\cubey}{2}
\pgfmathsetmacro{\cubez}{0.3}
\filldraw[draw=red, fill=red] (\span+2,0.2,-0.85) -- ++(-\cubex,0,0) -- ++(0,0,-\cubez) -- ++(\cubex,0,0) -- cycle;
\pgfmathsetmacro{\cubez}{2}
\draw[black] (\span+2,0.2,0) -- ++(-\cubex,0,0) -- ++(0,0,-\cubez) -- ++(\cubex,0,0) -- cycle;

\node (plus) at (\span + 1.2,-0.9,0) [above,font=\LARGE] {$+$} ;

\pgfmathsetmacro{\span}{9}
\pgfmathsetmacro{\cubex}{0.3}
\pgfmathsetmacro{\cubez}{2}
\pgfmathsetmacro{\cubey}{0.3}
\filldraw [draw=blue, fill=blue] (\span+2,-0.85) -- ++(-\cubex,0) -- ++(0,-\cubey) -- ++(\cubex,0) -- cycle;
\pgfmathsetmacro{\cubey}{2}
\draw[black] (\span + 2,0) -- ++(-\cubex,0) -- ++(0,-\cubey) -- ++(\cubex,0) -- cycle;

\node (A) at (\span+1.85,-2.5,0) [above] {$\bm{a}_D$} ;
\node (B) at (\span+3.2,-0.3,0) [below] {$\bm{b}_D$} ;
\node (C) at (\span+2.65,1.5,0) [below] {$\bm{c}_D$} ;
\node (i) at (\span+2.25,-0.7,0) [below] {$i$} ;
\node (j) at (\span+3.2,0.7,0) [below] {$j$} ;
\node (k) at (\span+1.85,1.0,0) [below] {$k$} ;

\pgfmathsetmacro{\cubey}{0.3}
\pgfmathsetmacro{\cubez}{2}
\pgfmathsetmacro{\cubex}{0.3}
\filldraw[draw=green, fill=green] (\span+3.35,0) -- ++(-\cubex,0) -- ++(0,-\cubey) -- ++(\cubex,0) -- cycle;
\pgfmathsetmacro{\cubex}{2}
\draw[black] (\span+4.2,0) -- ++(-\cubex,0) -- ++(0,-\cubey) -- ++(\cubex,0) -- cycle;

\pgfmathsetmacro{\cubex}{0.3}
\pgfmathsetmacro{\cubey}{2}
\pgfmathsetmacro{\cubez}{0.3}
\filldraw[draw=red, fill=red] (\span+2,0.2,-0.85) -- ++(-\cubex,0,0) -- ++(0,0,-\cubez) -- ++(\cubex,0,0) -- cycle;
\pgfmathsetmacro{\cubez}{2}
\draw[black] (\span+2,0.2,0) -- ++(-\cubex,0,0) -- ++(0,0,-\cubez) -- ++(\cubex,0,0) -- cycle;

%\pgfmathsetmacro{\cubex}{0.3}
%\pgfmathsetmacro{\cubey}{2}
%\pgfmathsetmacro{\cubez}{2}
%\draw[black] (\span + 2,0) -- ++(-\cubex,0) -- ++(0,-\cubey) -- ++(\cubex,0) -- cycle;
%\draw[black] (\span + 2,0) -- ++(0,0) -- ++(0,-\cubey) -- ++(0,0) -- cycle;
%\draw[black] (\span + 2,0) -- ++(-\cubex,0) -- ++(0,0) -- ++(\cubex,0) -- cycle;
%\node (A) at (\span + 1.85,-2.4,0) [above] {$\bm{a}$} ;
%\node (B) at (\span + 3.2,-0.3,0) [below] {$\bm{b}$} ;
%\node (C) at (\span + 2.65,1.4,0) [below] {$\bm{c}$} ;
%
%\pgfmathsetmacro{\cubex}{2}
%\pgfmathsetmacro{\cubey}{0.3}
%\pgfmathsetmacro{\cubez}{2}
%\draw[black] (\span + 4.2,0) -- ++(-\cubex,0) -- ++(0,-\cubey) -- ++(\cubex,0) -- cycle;
%\draw[black] (\span + 4.2,0) -- ++(0,0) -- ++(0,-\cubey) -- ++(0,0) -- cycle;
%\draw[black] (\span + 4.2,0) -- ++(-\cubex,0) -- ++(0,0) -- ++(\cubex,0) -- cycle;
%
%\pgfmathsetmacro{\cubex}{0.3}
%\pgfmathsetmacro{\cubey}{2}
%\pgfmathsetmacro{\cubez}{2}
%\draw[black] (\span + 2,0.2,0) -- ++(-\cubex,0,0) -- ++(0,0,0) -- ++(\cubex,0,0) -- cycle;
%\draw[black] (\span + 2,0.2,0) -- ++(0,0,-\cubez) -- ++(0,0,0) -- ++(0,0,\cubez) -- cycle;
%\draw[black] (\span + 2,0.2,0) -- ++(-\cubex,0,0) -- ++(0,0,-\cubez) -- ++(\cubex,0,0) -- cycle;
%
\node (plus) at (\span -1,-0.9,0) [above,font=\LARGE] {$+$} ;
\node (plus) at (\span +1.2,-0.9,0) [above,font=\LARGE] {$+$} ;
\node (dot) at (\span +0.1,-0.9,0) [above,font=\LARGE] {$\cdots$} ;

\end{tikzpicture}
\caption{Illustration of a $D$-component CP model for
a third-order tensor $\bm{\mathcal{X}}$.}
\label{fig:cp}
\end{figure}

%<<<<<<< HEAD
%CP decomposition~\cite{cp} is a way 
%to factorize a %third-order \memo{CP isn't restricted to 3rd order?}
%tensor $\bm{\mathcal{X}} \in \mathbb{R}^{N_e \times N_e \times N_r}$
%as a linear combination of $D$ rank-one tensors
%=======
CP decomposition~\cite{cp} factorizes a given tensor
as a linear combination of $D$ rank-one tensors.
%\memo{CP decomposition factorizes a tensor into a sum of component rank-one tensors.}
For a third-order
tensor $\bm{\mathcal{X}} \in \mathbb{R}^{N_e \times N_e \times N_r}$,
its CP decomposition is
%>>>>>>> 3bbe67bded5b90df7ee56926771a61bd08621460
\begin{displaymath}
  %% \bm{\mathcal{X}} \approx \sum_{d=1}^D \bm{a}_d \otimes \bm{b}_d \otimes \bm{c}_d
	\bm{\mathcal{X}} \approx \sum_{ d \in [D] } \bm{a}_d \otimes \bm{b}_d \otimes \bm{c}_d \text{~,}
\end{displaymath}
where $\bm{a}_d \in \mathbb{R}^{N_e}$, 
$\bm{b}_d \in \mathbb{R}^{N_e}$ and $\bm{c}_d \in \mathbb{R}^{N_r}$.
Figure~\ref{fig:cp} illustrates CP for third-order tensors,
which demonstrates how we can formulate knowledge graphs.
The elements $x_{ijk}$ of $\bm{\mathcal{X}}$ can be written as 
\begin{displaymath}
  %% x_{ijk} \approx \theta_{ijk} = \bm{a}_{i:}\transpose(\bm{b}_{j:}\circ\bm{c}_{k:})=\sum_{d=1}^D a_{id}b_{jd}c_{kd}
  x_{ijk} \approx \bm{a}_{i:}(\bm{b}_{j:}\circ\bm{c}_{k:})\transpose=\sum_{ d \in [D] }  a_{id}b_{jd}c_{kd}
  \quad \text{for $i,j \in [N_e],\ k\in[N_r]$.}
\end{displaymath}
%% \memo{
%% \begin{displaymath}
%% %%  x_{ijk} \approx \bm{a}_{i:}\transpose(\bm{b}_{j:}\circ\bm{c}_{k:})=\sum_{d=1}^D a_{id}b_{jd}c_{kd}
%%   x_{ijk} \approx \bm{a}_{i:}\transpose(\bm{b}_{j:}\circ\bm{c}_{k:})=\sum_{ d \in [D] } a_{id}b_{jd}c_{kd}
%%   \quad \text{for $i,j \in [N_e],\ k\in[N_r]$.}
%% \end{displaymath}
%% }
A \emph{factor matrix} refers to a matrix composed of vectors
%% The factor matrices refer to the combination \memo{is ``combination'' really the right terminology???}
from the rank one components.
We use 
$\bm{A}=[\bm{a}_1 \, \bm{a}_2 \, \cdots \, \bm{a}_D]$ to denote the 
%\memo{mode-oneを消したほうがいい}
factor matrix,
and likewise $\bm{B}$, $\bm{C}$.
%for the mode-two and -three factor matrices.
Note that
$\bm{a}_{i:}$, $\bm{b}_{j:}$ and $\bm{c}_{k:}$ represent the $D$-dimensional embedding vectors of subject $e_i$, object $e_j$,
and relation $r_k$, respectively.
%Further, $\theta_{ijk}$ can be regarded as the 
%score that represents the
%CP decomposition model's confidence that
%a triple~$(e_i,e_j,r_k)$ is a fact; i.e., that it must be present in the knowlege graph.
%\memo{further以下を削除}

\subsection{Logistic Regression}
\label{sec:logistic}
Following literature~\cite{logit}, we formulate a logistic regression model for
solving the CP decomposition problem.
This model considers 
CP decomposition from a probabilistic viewpoint.
We regard $x_{ijk}$ as a random variable and 
compute the maximum a posteriori~(MAP) estimates of $\bm{A}$, $\bm{B}$, and $\bm{C}$ for the joint distribution
\begin{displaymath}
p(\bm{\mathcal{X}}|\bm{A},\bm{B},\bm{C})=\prod_{ i \in [N_e] }
\prod_{ j \in [N_e] }\prod_{ k \in [N_r] } p(x_{ijk}|\theta_{ijk}).
%% p(\bm{\mathcal{X}}|\bm{A},\bm{B},\bm{C})=\prod_{i=1}^{N_e}
%% \prod_{j=1}^{N_e}\prod_{k=1}^{N_r} p(x_{ijk}|\theta_{ijk}).
\end{displaymath}
%\memo{
We define the score function 
$\theta_{ijk} = \bm{a}_{i:}(\bm{b}_{j:}\circ\bm{c}_{k:})\transpose$
that represents the 
CP decomposition model's confidence that a triple $(e_i,e_j,r_k)$ is a fact;
i.e., that it must be present in the knowledge graph. 
%}
By assuming that $x_{ijk}$ follows the Bernoulli distribution,
$
x_{ijk} \sim \text{Bernoulli} (\sigma(\theta_{ijk}))
$,
the posterior probability is defined as the following equation
\begin{displaymath}
p(x_{ijk}|\theta_{ijk}) = \left\{
\begin{array}{ll}
\sigma(\theta_{ijk}) & \quad \text{if } x_{ijk}=1 \\
1 - \sigma(\theta_{ijk}) & \quad \text{if } x_{ijk}=0
\end{array} \right.
,
\end{displaymath}
where ~$\sigma(x)=1/ ( 1 + \exp(-x) )$ is the sigmoid function.

Furthermore, we minimize the negative log-likelihood of the MAP estimates,
such that the general form of the objective function to optimize is

%\begin{displaymath}
%\argmin_{\bm{A},\bm{B},\bm{C}} \bigl\{\text{loss}(\bm{\mathcal{X}};\bm{A},\bm{B},\bm{C})
%+ \lambda_A \|\bm{A}\|_F^2 + \lambda_B \|\bm{B}\|_F^2 + \lambda_C \|\bm{C}\|_F^2\ \bigr\}.
%\end{displaymath}
%\memo{
%\begin{displaymath}
%E = \text{loss}(\bm{\mathcal{X}};\bm{A},\bm{B},\bm{C})
%+ \lambda_A \|\bm{A}\|_F^2 + \lambda_B \|\bm{B}\|_F^2 + \lambda_C \|\bm{C}\|_F^2,
%\end{displaymath}
%}
%\memo{
\begin{displaymath}
  %% E = \sum_{i=1}^{N_e}\sum_{j=1}^{N_e}\sum_{k=1}^{N_r}E^{(ijk)},
  E = \sum_{ i \in [N_e] }\sum_{ j \in [N_e] }\sum_{ k \in [N_r] }E_{ijk},
\end{displaymath}
where
%\begin{displaymath}
%E^{(ijk)} = \text{loss}(\bm{\mathcal{X}};\bm{A},\bm{B},\bm{C})
%+ \lambda_A \|\bm{A}\|_F^2 + \lambda_B \|\bm{B}\|_F^2 + \lambda_C \|\bm{C}\|_F^2,
%\end{displaymath}
\begin{eqnarray*}
\begin{split}
	E_{ijk} &= 
	\underbrace{-x_{ijk}\log{\sigma(\theta_{ijk}}) + (x_{ijk}-1) \log(1- \sigma(\theta_{ijk}))}_{
\let\scriptstyle\textstyle
	\substack{\ell_{ijk}}} \\
	&\qquad \qquad \qquad \qquad \qquad + \underbrace{ \lambda_A\|\bm{a}_{i:}\|^2 
	+ \lambda_B\|\bm{b}_{j:}\|^2 
	+ \lambda_C\|\bm{c}_{k:}\|^2 }_{
\let\scriptstyle\textstyle
	\substack{\text{L2 regularizer}}}.
\end{split}
\end{eqnarray*}$\ell_{ijk}$ represents the logistic loss function for
a triple $(e_i,e_j,r_k)$.
While
%}
%where the loss function is defined as
%\begin{displaymath}
%  \text{loss}(\bm{\mathcal{X}};\bm{A},\bm{B},\bm{C}) = %\coloneqq
%  - \sum_{i=1}^{N_e}
%\sum_{j=1}^{N_e}\sum_{k=1}^{N_r}
%\bigl\{
%x_{ijk}\log{\sigma(\theta_{ijk}}) + (1-x_{ijk}) \log(1- \sigma(\theta_{ijk}))\bigr\}.
%\end{displaymath}
most knowledge graphs contain only positive examples,
negative examples (false facts) are needed to optimize the objective function.
However, if all unknown triples are treated as negative samples,
calculating the loss function requires
a prohibitive amount of time.
To approximately minimize the objective function,
following previous studies,
we used negative sampling in our experiments.

The objective function is minimized with an online learning method based on stochastic gradient descent~(SGD).
%% SGD-based online-learning methods are often
%% used to solve large-scale optimization problems.
For each training example, SGD iteratively updates parameters by
$\bm{a}_{i:} \leftarrow \bm{a}_{i:}-\eta\frac{\partial E_{ijk}}{\partial \bm{a}_{i:}}$,
$\bm{b}_{j:} \leftarrow \bm{b}_{j:}-\eta\frac{\partial E_{ijk}}{\partial \bm{b}_{j:}}$,
and
$\bm{c}_{k:} \leftarrow \bm{c}_{k:}-\eta\frac{\partial E_{ijk}}{\partial \bm{c}_{k:}}$
with a learning rate $\eta$.
The partial gradient of the objective function with respect to
$\bm{a}_{i:}$ is
\begin{eqnarray*}
	\frac{\partial E_{ijk}}{\partial \bm{a}_{i:}}
&=&
-x_{ijk}
%\frac{\exp{\left(-\theta_{ijk} \right)}}
%{1 + \exp{\left(-\theta_{ijk} \right)}}
\exp{\left(-\theta_{ijk} \right)}\sigma(\theta_{ijk})
\bm{b}_{j:} \circ \bm{c}_{k:}
+ \left(1-x_{ijk}\right)
\sigma(\theta_{ijk})
%\frac{1}{1 + \exp{\left(-\theta_{ijk} \right)}}
\bm{b}_{j:} \circ \bm{c}_{k:}
+2\lambda_A \bm{a}_{i:}.
\end{eqnarray*}
Those with respect to $\bm{b}_{j:}$ and
$\bm{c}_{k:}$ can be calculated in the same manner.

%%% Local Variables:
%%% mode: latex
%%% TeX-PDF-mode: t
%%% TeX-engine: xetex
%%% TeX-master: "main"
%%% End:

\section{Binarized CP Decomposition}
\label{sec:prop}
We propose a binarized CP decomposition algorithm
to make CP factor matrices
$\bm{A}$, $\bm{B}$ and $\bm{C}$
binary, i.e., the elements of these matrices are constrained to 
only two possible values.

In this algorithm, we formulate the score function 
%% $\theta_{ijk}^{(b)}=\sum_{d=1}^Da_{id}^{(b)}b_{jd}^{(b)}c_{kd}^{(b)}$
$\theta_{i j k}^{(b)}=\sum_{ d \in [D] } a_{i d}^{(b)}b_{j d}^{(b)}c_{k d}^{(b)}$,
where $a_{i d}^{(b)}=Q_{\Delta}(a_{i d}),\ b_{j d}^{(b)}=Q_{\Delta}(b_{j d}),
\ c_{k d}^{(b)}=Q_{\Delta}(c_{k d})$ are 
obtained by binarizing 
$a_{i d},\ b_{j d},\ c_{k d}$ through
the following quantization function
\begin{displaymath}
x^{(b)}=Q_{\Delta}(x)=
\begin{cases} 
   \Delta  & \quad \text{if } x \ge 0 \\
   -\Delta & \quad \text{if } x < 0
\end{cases},
\end{displaymath}
where
$\Delta$
is a positive constant value.
We extend the binarization function to
vectors in a natural way:
$\bm{x}^{(b)}=Q_{\Delta}(\bm{x})$
whose $i$th element $x_i^{(b)}$ is $Q_{\Delta}(x_i)$.

Using the new score function,
we reformulate the loss function defined in Section \ref{sec:logistic}
as follows
%\begin{displaymath}
%  \text{loss}_{b}(\bm{\mathcal{X}};\bm{A},\bm{B},\bm{C}) = % \coloneqq
%  - 
%\sum_{i=1}^{N_e}\sum_{j=1}^{N_e}\sum_{k=1}^{N_k}
%x_{ijk}\log{\sigma(\theta_{ijk}^{(b)}}) + (1-x_{ijk}) \log(1- \sigma(\theta_{ijk}^{(b)})).
%\end{displaymath}
\begin{displaymath}
	\ell_{ijk}^{(b)}= 
	-x_{ijk}\log{\sigma(\theta_{ijk}^{(b)})} 
	+ (x_{ijk}-1) \log(1- \sigma(\theta_{ijk}^{(b)})).
\end{displaymath}
To train the binarized CP decomposition model,
we optimize the same objective function $E$
as in Section \ref{sec:logistic}
except using the binarized loss function given above.
We also employ the SGD algorithm to minimize the objective function.
One issue here is that the parameters cannot be updated properly
since the gradients of $Q_{\Delta}$ are zero
almost everywhere.
To solve the issue, we simply
use an identity matrix as the surrogate for the derivative
of $Q_{\Delta}$:
\begin{displaymath}
\frac{\partial Q_{\Delta}(\bm{x})}{\partial \bm{x}} \approx \bm{I}.
\end{displaymath}
%where $\bm{j}$ is a vector where every element is one.
The simple trick enables us to calculate the partial gradient of
the objective function with respect to
$\bm{a}_{i:}$ through the chain rule:
\begin{displaymath}
	\frac{\partial \ell^{(b)}_{ijk}}{\partial \bm{a}_{i:}}=
\frac{\partial Q_{\Delta}(\bm{a}_{i:})}{\partial \bm{a}_{i:}} 
\frac{\partial \ell^{(b)}_{ijk}}{\partial Q_{\Delta}(\bm{a}_{i:})} \approx
\bm{I}
\frac{\partial \ell^{(b)}_{ijk}}{\partial Q_{\Delta}(\bm{a}_{i:})} =
\frac{\partial \ell^{(b)}_{ijk}}{\partial \bm{a}_{i:}^{(b)}}.
\end{displaymath}
This strategy is known as Hinton's straight-through estimator~\cite{hinton}
and has been developed in the community of 
deep neural networks
to quantize network components~\cite{bicon,surveyq}.
Using this trick,
we finally obtain the partial gradient as follows:
\begin{eqnarray*}
	\frac{\partial E_{ijk}}{\partial \bm{a}_{i:}}
&=&
-x_{ijk}
%\frac{\exp{\left(-\theta_{ijk}^{(b)} \right)}}{1 + \exp{\left(-\theta_{ijk}^{(b)} \right)}}
\exp{\left(-\theta_{ijk}^{(b)} \right)}\sigma(\theta_{ijk}^{(b)})
\bm{b}_{j:}^{(b)} \circ \bm{c}_{k:}^{(b)}
+ \left(1-x_{ijk}\right)
%\frac{1}{1 + \exp{\left(-\theta_{ijk}^{(b)} \right)}}
\sigma(\theta_{ijk}^{(b)})
\bm{b}_{j:}^{(b)} \circ \bm{c}_{k:}^{(b)}
+2\lambda_A \bm{a}_{i:}.
\end{eqnarray*}
The partial gradients with respect to $\bm{b}_{j:}$ and $\bm{c}_{k:}$ 
can be computed similarly.

%Adopting this method, we can not 
%only save memory but also
%improve computing speed.
%In the case of calculating the score function,
%we can compute fast by replacing innner product of 
%floating point value with 
%counting the true value of Xnor. 
%The output value of Xnor is inverse of exclusive disjunction.
%We discuss the method to calculate 
%the score function which
%is computed as
%%As mentioned in Section~\ref{sec:intro},
Binary vector representations
bring benefits in
faster computation of scores $\theta_{ijk}^{(b)}$,
because
the inner product between binary vectors can
be implemented by bitwise operations:
To compute $\theta_{ijk}^{(b)}$,
we can use XNOR and Bitcount operations:
\begin{displaymath}
	\theta_{ijk}^{(b)}=\bm{a}_{i:}^{(b)}(\bm{b}_{j:}^{(b)}\circ
\bm{c}_{k:}^{(b)})\transpose=\Delta^{3} \{
2 BitC-D
\}
\end{displaymath}
where $BitC=\text{Bitcount}(\text{XNOR}(\text{XNOR}(\overline{\bm{a}}_{i:}^{(b)},\overline{\bm{b}}_{j:}^{(b)}),\overline{\bm{c}}_{k:}^{(b)}))$.
$\overline{\bm{x}}^{(b)}$ denotes the boolean vector 
%obtained by converting $x^{(b)}_i$ to 1
%if $x_i^{(b)}$ is $\Delta$, otherwise to 0.
whose $i$th element $\overline{x}_i^{(b)}$ is set to 1 if $x_i^{(b)}=\Delta$,
otherwise to 0.
Bitcount returns the number of one-bits in a binary vector
and $\text{XNOR}$ represents
the logical complement of the exclusive OR operation.

%%%%%%%%%%%%%%%%%%%%%%%%%%%%%%%%%%%%%%%%%%%%%%%%%%%%%%%%%%%%%%%%%%%%%%%%

%%% LocalWords: KGC KGE CANDECOMP PARAFAC CP SGD
%%% LocalWords: ComplEx HolE SimplE NTN TransE TransH STransE TransR ConvE
%%% LocalWords: bilinear overfitting nonsymmetric binarize binarized binarizing binarization quantize quantized
%%% LocalWords: Hayashi Shimbo Hermitian Kazemi Lacroix et al th
%%% LocalWords: MNIST CIFAR BinaryConnect XNOR Bitcount BitC boolean
%%% LocalWords: ijk

%%% Local Variables:
%%% mode: latex
%%% TeX-PDF-mode: t
%%% TeX-engine: xetex
%%% TeX-master: "main"
%%% End:

%\section{Related Work}
%\input{related}

\section{Experiments}
\label{sec:exp}
\subsection{Datasets and Evaluation Protocol}
\begin{table}[t]
  \centering
  %\small
  \mytablefont
  \caption{Benchmark datasets for KGC.}
  \label{tab:dataset}
  \begin{tabular}{lrrrr}
  \toprule
                             & WN18 & FB15k & WN18RR & FB15k-237 \\\midrule
             $N_e$ & 40,943 & 14,951 & 40,559 & 14,505 \\
             $N_r$ & 18 & 1,345 & 11 & 237 \\
             \# training triples & 141,442 & 483,142 & 86,835 & 272,115 \\
             \# validation triples & 5,000 & 50,000 & 3,034 & 17,535 \\
             \# test triples  & 5,000 & 59,071 & 3,134 & 20,466 \\
  \bottomrule
  \end{tabular} 
  %\scalebox{1.0}{
  %\begin{tabular}{lrrrrrrrrr}
  %  \toprule
  %  & & &        & Base    &        &         & \multicolumn{2}{c}{Path}        & \\
  %  & $|\mathcal{E}|$ & $|\mathcal{R}|$ & \#train & \#valid & \#test & \#train & \#valid   & \#test & \#test-uniq \\
  %  % \cmidrule(lr){1-1}\cmidrule(lr){2-2}\cmidrule(lr){3-3}\cmidrule(lr){4-4}\cmidrule(lr){5-5}\cmidrule(lr){6-6}
  %  \cmidrule(lr){1-10}
  %  WN18  & & & 141,442 & 5,000 & 5,000 \\%& 5,760,450 & 10,000 & 10,000 & 13,133 \\
  %  FB15k & & & 483,142 & 50,000 & 59,071 \\%& 7,861,321 & 30,000 & 30,000 & 59,586 \\
  %  YAGO3-10 & \\
  %  \bottomrule
  %\end{tabular}
  %}
\end{table}

%%%%%%%%%%%%%%%%%%%%%%%%%%%%%%%%%%%%%%%%%%%%%%%%%%%%%%%%%%%%%%%%%%%%%%%% 

%% Local Variables:
%% Mode: LaTeX
%% TeX-master: "../main"
%% TeX-engine: xetex
%% TeX-PDF-mode: t
%% End:

\begin{table*}[tb]
  %\scriptsize
  \mytablefont
  \caption{KGC results on WN18 and FB15k: Filtered MRR and Hits@$\{1,3,10\}$ (\%).
    Letters in boldface signify the best performers in individual evaluation metrics.
      *, ** and *** indicate the results transcribed from~\cite{complex},~\cite{conve} and~\cite{simple}, respectively.}
  \label{tab:results_old}
  \centering
    \begin{tabular}{lcccccccc}
      \toprule
      & \multicolumn{4}{c}{WN18}   & \multicolumn{4}{c}{FB15k} \\
      \cmidrule(lr){2-5}\cmidrule(lr){6-9}
      & \multirow{2}{*}{MRR} & \multicolumn{3}{c}{Hits@} & \multirow{2}{*}{MRR} & \multicolumn{3}{c}{Hits@} \\
      \cmidrule(lr){3-5}\cmidrule(lr){7-9}
      Models & & 1 & 3 & 10 & & 1 & 3 & 10  \\\midrule
      TransE*  & 45.4 & 8.9  & 82.3 & 93.4 & 38.0 & 23.1 & 47.2 & 64.1  \\
%      RESCAL** & 89.0 & 60.3 & 84.2 & 90.4 & 92.8 & 35.4 & 18.9 & 23.5 & 40.9 & 58.7  \\
      DistMult*& 82.2 & 72.8 & 91.4 & 93.6 & 65.4 & 54.6 & 73.3 & 82.4  \\
      HolE*    & 93.8 & 93.0 & 94.5 & 94.9 & 52.4 & 40.2 & 61.3 & 73.9  \\
      ComplEx* & 94.1 & 93.6 & 94.5 & 94.7 & 69.2 & 59.9 & 75.9 & 84.0  \\
      ANALOGY**& 94.2 & 93.9 & 94.4 & 94.7 & 72.5 & 64.6 & {\bf 78.5} & {\bf 85.4}  \\
      CP***    & 94.2 & 93.9 & 94.4 & 94.7 & 72.7 & 66.0 & 77.3 & 83.9 \\
      ConvE**  & 94.3 & 93.5 & 94.6 & {\bf 95.6} & 65.7 & 55.8 & 72.3 & 83.1  \\\midrule
      CP ($D=200$)      & 94.2 & 93.9 & 94.5 & 94.7 & 71.9 & 66.2 & 75.2 & 82.0 \\
      {\bf B-CP} ($D=200$) & 90.1 & 88.1 & 91.8 & 93.3 & 69.5 & 61.1 & 76.0 & 83.5 \\
      {\bf B-CP} ($D=400$) & 94.5 & 94.1 & 94.8 & 95.0 & 72.2 & 66.3 & 77.5 & 84.2 \\
      {\bf B-CP} ($D=300\times3$) & {\bf 94.6} & {\bf 94.2} & {\bf 95.0} & 95.3 & {\bf 72.9} & {\bf 66.5} & 77.7 & 84.9 \\\bottomrule
      %{\bf B-CP} ($D=400\times3$) \\\bottomrule
    \end{tabular}
\end{table*}

%%%%%%%%%%%%%%%%%%%%%%%%%%%%%%%%%%%%%%%%%%%%%%%%%%%%%%%%%%%%%%%%%%%%%%%% 

%% Local Variables:
%% Mode: LaTeX
%% TeX-master: "../main"
%% TeX-engine: xetex
%% TeX-PDF-mode: t
%% End:

\begin{table*}[tb]
  \mytablefont
  \caption{KGC results on WN18RR and FB15k-237: Filtered MRR and Hits@$\{1,3,10\}$ (\%).
      * indicates the results transcribed from~\cite{conve}.}
  \label{tab:results_new}
  \centering
    \begin{tabular}{lcccccccc}
      \toprule
      & \multicolumn{4}{c}{WN18RR}   & \multicolumn{4}{c}{FB15k-237} \\
      \cmidrule(lr){2-5}\cmidrule(lr){6-9}
      & \multirow{2}{*}{MRR} & \multicolumn{3}{c}{Hits@} & \multirow{2}{*}{MRR} & \multicolumn{3}{c}{Hits@} \\
      \cmidrule(lr){3-5}\cmidrule(lr){7-9}
      Models    &      & 1    & 3    & 10   &      & 1    & 3    & 10  \\\midrule
      DistMult* & 43.0 & 39.0 & 44.0 & 49.0 & 24.1 & 15.5 & 26.3 & 41.9 \\
      ComplEx*  & 44.0 & 41.0 & 46.0 & 51.0 & 24.7 & 15.8 & 27.5 & 42.8 \\
      R-GCN*    & --   & --   & --   & --   & 24.8 & 15.3 & 25.8 & 41.7 \\
      ConvE*    & 43.0 & 40.0 & 44.0 & 52.0 & {\bf 32.5} & {\bf 23.7} & {\bf 35.6} & {\bf 50.1} \\\midrule
      CP ($D=200$)       & 44.0 & 42.0 & 46.0 & 51.0 & 29.0 & 19.8 & 32.2 & 47.9 \\
      {\bf B-CP} ($D=200$) & 45.0 & 43.0 & 46.0 & 50.0 & 27.8 & 19.4 & 30.4 & 44.6 \\
      {\bf B-CP} ($D=400$) & 45.0 & 43.0 & 46.0 & 52.0 & 29.2 & 20.8 & 31.8 & 46.1 \\
      {\bf B-CP} ($D=300\times3$) & {\bf 48.0} & {\bf 45.0} & {\bf 49.0} & {\bf 53.0} & 30.3 & 21.4 & 33.3 & 48.2 \\\bottomrule
      %{\bf B-CP} ($D=400\times3$) \\\bottomrule
    \end{tabular}
\end{table*}

%%%%%%%%%%%%%%%%%%%%%%%%%%%%%%%%%%%%%%%%%%%%%%%%%%%%%%%%%%%%%%%%%%%%%%%% 

%% Local Variables:
%% Mode: LaTeX
%% TeX-master: "../main"
%% TeX-engine: xetex
%% TeX-PDF-mode: t
%% End:

\begin{figure}[t]
\centering
\begin{tabular}{c}
\begin{minipage}{0.5\hsize}
\centering
\includegraphics[width=0.95\linewidth]{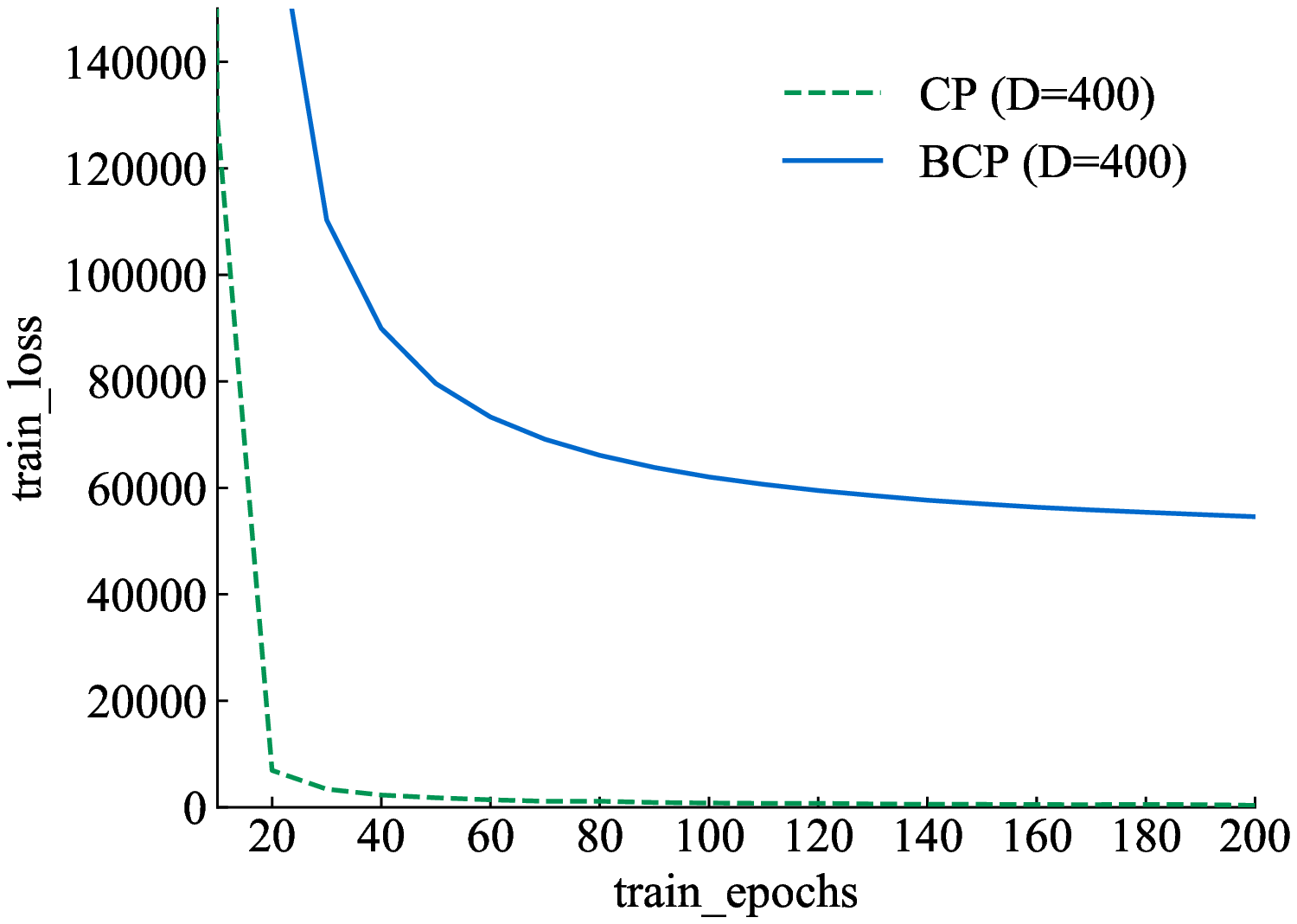}
\end{minipage}
\begin{minipage}{0.5\hsize}
\centering
\includegraphics[width=0.95\linewidth]{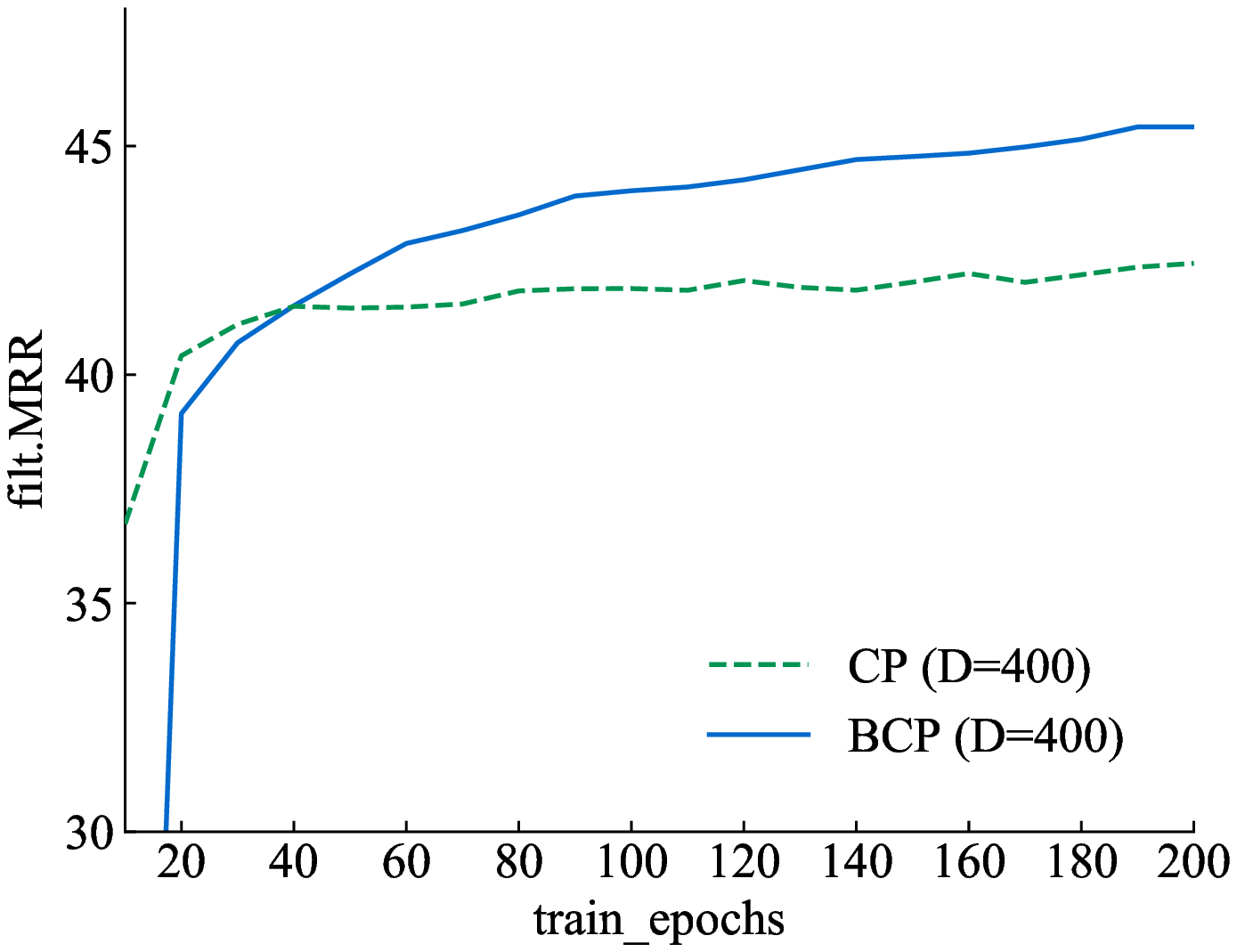}
\end{minipage}
\end{tabular}
% \caption{$bn=200$に固定して、Block HolEのブロック数$b$を2、4、8と変化させた時のWN18上でのランキング精度}
%\caption{Path QA classification result comparing BlockHolE ($b=2,n=50$ and $b=2,n=100$)
%to DistMult, ComplEx and RESCAL models.
%The dimensionality of the DistMult, ComplEx and RESCAL embeddings
%was set to 100.}
\caption{
Training loss and filtered MRR vs. epochs
trained on WN18RR.}
\label{fig:loss-acc}
\end{figure}

%%%%%%%%%%%%%%%%%%%%%%%%%%%%%%%%%%%%%%%%%%%%%%%%%%%%%%%%%%%%%%%%%%%%%%%% 

%% Local Variables:
%% Mode: LaTeX
%% TeX-master: "../main"
%% TeX-engine: xetex
%% TeX-PDF-mode: t
%% End:

\begin{table*}[tb]
  \mytablefont
  \caption{Results on WN18RR and FB15k-237 with varying embedding dimensions.}
  \label{tab:results_comp}
  \centering
    \begin{tabular}{lrrcc}
       \toprule
       \multirow{2}{*}{Model} & \multirow{2}{*}{Bits per entity} & \multirow{2}{*}{Bits per relation} & \multicolumn{2}{c}{MRR} \\
       & & & WN18RR & FB15k-237 \\\midrule
       DistMult* ($D=200$) & 6,400 & 6,400 & 43.0 & 24.1 \\
       ComplEx* ($D=200$) & 12,800 & 12,800 & 44.0 & 24.7 \\
       ConvE* ($D=200$)  & 6,400 & 6,400 & 43.0 & {\bf 32.5} \\\midrule
       CP ($D=15$)   & 960 & 480 & 40.0 & 22.0 \\
       CP ($D=50$)   & 3,200 & 1,600 & 43.0 & 24.8 \\
       CP ($D=200$)  & 12,800 & 6,400 & 44.0 & 29.0 \\
       CP ($D=500$)  & 32,000 & 16,000 & 43.0 & 29.2 \\
       %CP ($D=300\times3$)  & 57,600 & 28,800 &  &  \\ 
       VQ-CP ($D=200$) & 400 & 200 & 36.0 & 8.7 \\
       VQ-CP ($D=500$)  & 1,000 & 500 & 36.0 & 8.3 \\\midrule
       {\bf B-CP} ($D=100$) & 200 & 100 & 38.0 & 23.2 \\
       {\bf B-CP} ($D=200$) & 400 & 200 & 45.0 & 27.8 \\
       {\bf B-CP} ($D=300$) & 600 & 300 & 46.0 & 29.0 \\
       {\bf B-CP} ($D=400$) & 800 & 400 & 45.0 & 29.2 \\
       {\bf B-CP} ($D=500$) & 1,000 & 500 & 45.0 & 29.1 \\
       {\bf B-CP} ($D=300\times3$) & 1,800 & 900 & {\bf 48.0} & 30.3 \\\bottomrule
    \end{tabular}
\end{table*}

%%%%%%%%%%%%%%%%%%%%%%%%%%%%%%%%%%%%%%%%%%%%%%%%%%%%%%%%%%%%%%%%%%%%%%%% 

%% Local Variables:
%% Mode: LaTeX
%% TeX-master: "../main"
%% TeX-engine: xetex
%% TeX-PDF-mode: t
%% End:

\begin{figure}[t]
\centering
%\begin{tabular}{cc}
%\begin{minipage}{0.48\hsize}
%\centering
%\includegraphics[width=8cm,height=5cm]{figures/acc-wn18.eps}
%\caption{KGC result on WN18 with varying block sizes $b\in \{2, 4, 8\}$ while fixing $bn=200$ in BlockHolE.}
%\label{fig:acc}
%\end{minipage}
%&
%\begin{minipage}{0.48\hsize}
%\centering
%\includegraphics[width=13cm,height=7cm]{figures/result_float}
\includegraphics[height=0.3\textheight]{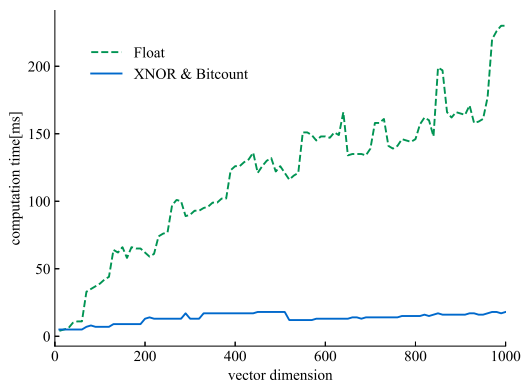}
\caption{CPU run time per 100,000-times score computations with single CPU thread.}
\label{fig:time}
%\end{minipage}
%\end{tabular}
\end{figure}

%%%%%%%%%%%%%%%%%%%%%%%%%%%%%%%%%%%%%%%%%%%%%%%%%%%%%%%%%%%%%%%%%%%%%%%% 

%%% Local Variables:
%%% Mode: LaTeX
%%% TeX-master: "../main"
%%% TeX-engine: xetex
%%% TeX-PDF-mode: t
%%% End:

We evaluated the performance of our proposal in the standard
knowledge graph completion (KGC) task.
We used four standard datasets,
WN18, FB15k~\cite{transe}, WN18RR, and FB15k-237~\cite{conve}.
Table~\ref{tab:dataset} shows the data statistics\footnote{
Following \cite{simple,cano},
  %% improve the KGC performance of CP models,
for each triple $(e_i,e_j,r_k)$ observed in the training set,
we added its inverse triple $(e_j,e_i,r_k^{-1})$
also in the training set.}.

We followed the standard evaluation procedure
to evaluate the KGC performance:
Given a test triple $(e_i,e_j,r_k)$, we
corrupted it by replacing $e_i$ or $e_j$
with every entity $e_\ell$ in $\mathcal{E}$
and calculated $\theta_{i,\ell,k}$ or $\theta_{\ell,j,k}$.
We then ranked all these triples by
their scores in decreasing order.
To measure the quality of the ranking,
we used the mean reciprocal rank (MRR) and Hits at $N$~(Hits@$N$).
%We here reported results
%in both the filtered and raw settings~\cite{transe}
%for MRR, but only filtered values for Hits@$N$.
We here report only results in the filtered setting~\cite{transe},
which provides a more reliable performance metric in the presence of multiple correct triples.

\subsection{Experiment Setup}
\label{sec:expcond}
To train CP models, we selected the hyperparameters via
grid search such that the filtered MRR is maximized on the validation set.
For standard CP model,
we tried all combinations of $\lambda_A,\lambda_B,\lambda_C\in\{0.0001, 0\}$,
learning rate $\eta\in\{0.025, 0.05\}$,
and the embedding dimension $D\in\{15, 25, 50, 100, 150, \allowbreak 200, \allowbreak 300, \allowbreak 400, \allowbreak 500\}$ during grid search.
For our binarized CP (B-CP) model,
all combinations of $\lambda_A,\lambda_B,\lambda_C\in\{0.0001, 0\}$,
$\eta\in\{0.025, 0.05\}$, $\Delta\in\{0.3, 0.5\}$
and $D\in\{100,200,300,400,500\}$ were tried.
%\memo{The ensemble model isn't mentioned here at all. Its explanation must be placed here ore somewhere else, including how it aggregates the scores from 3 models.}
We randomly generated
the initial values of the representation vectors
from the uniform distribution
$U[-\frac{\sqrt{6}}{\sqrt{2D}},\frac{\sqrt{6}}{\sqrt{2D}}]$~\cite{init}.
The maximum number of training epochs was set to 1000.
For SGD training, negative samples were generated using
the local closed-world assumption~\cite{transe}.
The number of negative samples
generated per positive sample was 5 for WN18/WN18RR
and 10 for FB15k/FB15k-237.
%\memo{When we evaluated B-CP model, we also used {\it model ensemble}, which is the method
%using the sum of the scores in separately trained models.}
%\textcolor{red}{
%	%Since  B-CP model has less memory consumption
%%compared to CP model, 
%For improving the task performance,
%we used {\it model ensemble} which combines 
%the predictions from multiple models, %for improving the task performance.
%%we tried to combine the predictions from multiple B-CP models for improving the task performance.
%%Model ensemble is the method to use total scores of separately trained models,
%however,
%Model ensemble increases memory footprint in proportion to the number of trained models.}
%When we evaluated B-CP model, we also used {\it model ensemble} which sums up
%the scores in separately trained models.}
%\textcolor{red}{
In addition, to further take advantage of the benign run-time memory footprint of B-CP,
we also tested the {\it model ensemble} of three independently trained B-CP models\footnote{
  As the original CP model has much larger memory consumption than B-CP, we did not 
  test model ensemble with the CP model in our experiments.},
in which the final ranking is computed by the sum of the scores of the three models.
For this ensemble, the embedding dimension of each model was set to $D=300$,
yet the total required run-time memory is still smaller than CP with $D=200$.
%Model ensemble is the method to use total scores of separately trained models
%,which consumes memory in proprtion to the number of trained models.}

We implemented our CP decomposition 
systems in C++ and conducted all experiments on a 64-bit
16-Core AMD Ryzen Threadripper 1950x with 3.4GHz CPUs.
The program codes were compiled using GCC 7.3 with -O3 option.

\subsection{Results}
\subsubsection{Main Results}
We compared standard CP and 
%binarized CP models~(B-CP)
%binarized CP (B-CP) 
B-CP models
with other state-of-the-art KGE models.
Table~\ref{tab:results_old} shows
the results on WN18 and FB15k,
and Table~\ref{tab:results_new} displays the results
on WN18RR and FB15k-237.
For most of the evaluation metrics,
our B-CP model~($D=400$) outperformed or was competitive
to the best baseline, although
with a small vector dimension ($D=200$),
B-CP showed tendency to degrade in its performance.
In the table, B-CP $(D=300\times 3)$ indicates
an ensemble of three different B-CP models (each with $D=300$).
This ensemble approach outperformed the baseline B-CP
constantly on all datasets.
Figure~\ref{fig:loss-acc} shows
training loss and accuracy versus epochs of training
for CP~($D=400$) and B-CP~($D=400$) on WN18RR.
The results indicate that CP is prone to overfitting with increased
epochs of training.
By contrast, B-CP appears less susceptible to
overfitting than CP.
%The results show that
%the quantization acts as regularizer.
%\memo{The data included in a knowledge graph is few compared with
%all possible triples, and therefore 
%the conventional CP decomposition method overfit to training data.}

\subsubsection{KGC Performance vs. Model Size}
We also investigated how our B-CP method
can maintain the KGC performance while reducing the model size.
For a fair evaluation, we also examined a naive vector quantization
method (VQ)~\cite{xnor} that can reduce the model size.
Given a real valued matrix ${\bm X}\in\Rset^{D_1\times D_2}$,
the VQ method solves the following optimization problem:
\begin{align*}
{\hat {\bm X}^{(b)}},{\hat \alpha}=\argmin_{{\bm X}^{(b)},\alpha}
{\|{\bm X}-\alpha{\bm X}^{(b)}\|_F^2}
\end{align*}
where ${\bm X}^{(b)}\in\{+1,-1\}^{D_1\times D_2}$ is a binary matrix
and $\alpha$ is a positive real value.
The optimal solutions ${\hat {\bm X}^{(b)}}$ and ${\hat \alpha}$
are given by $Q_1({\bm X})$ and $\frac{1}{D_1\times D_2} \| {\bm X} \|_{1}$,
respectively,
where $\|\cdot \|_1$ denotes $l_1$-norm,
and
$Q_1({\bm X})$ is a sign function whose behavior in each element $x$
of ${\bm X}$ is as per the sign function $Q_1(x)$.
%\begin{align*}
%\text{sign}(x)=\begin{cases}
%    1 & x\geq 0 \\
%    -1 & \text{otherwise}
%\end{cases}.
%\end{align*}
After obtaining factor matrices ${\bm A}$, ${\bm B}$ and ${\bm C}$
via CP decomposition,
we solved the above optimization problem independently for each matrix.
We call this method VQ-CP.

Table~\ref{tab:results_comp} shows the results
when the dimension size of the embeddings was varied.
While CP requires $64\times D$ and $32\times D$ bits per
entity and relation, respectively,
both B-CP and VQ-CP have only to take one thirty-second of them.
Obviously, the task performance dropped significantly
after vector quantization~(VQ-CP).
The performance of CP also degraded when reducing
the vector dimension from 200 to 15 or 50.
While simply reducing the number of dimensions degraded
the accuracy,
B-CP successfully reduced the
model size nearly 10- to 20-fold compared to CP and other KGE models
without performance degradation.
Even in the case of B-CP~$(D=300\times3)$, 
the model size was 6 times smaller than that of CP~$(D=200)$.

\subsubsection{Computation Time}
%While B-CP requires keeping the real-valued weights during training to do SGD updates,
%we can maintain just binary weights in testing time.
%The B-CP model contributes to not only model size reduction
As described in Section~\ref{sec:prop}, the B-CP model can
accelerate the computation of confidence scores by using the
bitwise operations~(XNOR and Bitcount).
%since
%the inner product between binary weight vectors
%can be efficiently implemented by the bitwise~(Xnor and Bitcount) operations.
To compare the score computation speed between CP~(Float) and
B-CP~(XNOR and Bitcount),
we calculated the confidence scores 100,000 times for both CP and B-CP,
varying the vector size $D$ from 10 to 1000 at 10 increments.
Figure~\ref{fig:time} clearly shows that
bitwise operations provide significant speed-up compared to
standard multiply-accumulate operations.
%The fast score computation would also bring great benefits
%to model training time.
%Actually, B-CP~$(D=500)$ was almost as fast as CP~$(D=200)$ over all
%datasets\footnote{

%Since B-CP needs to keep and binarize real-valued vectors
%during SGD updates, however,
%it took about twice as long per training epoch as CP.
%We suspect this is mainly because the GCC optimizer did not work well for
%the quantization process.

\subsection{Evaluation on Large-scale Freebase}
\begin{table}[t]
  \centering
  \mytablefont
  \caption{Results on the Freebase-music dataset.}
  \label{tab:ldata}
  \begin{tabular}{lcccc}
  \toprule
                 & Accuracy & Model size \\\midrule
  CP~$(D=15)$    & 50.3         & 0.4GB \\
  CP~$(D=200)$   & 89.2     & 4.8GB \\
  {\bf B-CP}~$(D=400)$ & {\bf 92.8}     & {\bf 0.3GB} \\
  \bottomrule
  \end{tabular} 
  %\scalebox{1.0}{
  %\begin{tabular}{lrrrrrrrrr}
  %  \toprule
  %  & & &        & Base    &        &         & \multicolumn{2}{c}{Path}        & \\
  %  & $|\mathcal{E}|$ & $|\mathcal{R}|$ & \#train & \#valid & \#test & \#train & \#valid   & \#test & \#test-uniq \\
  %  % \cmidrule(lr){1-1}\cmidrule(lr){2-2}\cmidrule(lr){3-3}\cmidrule(lr){4-4}\cmidrule(lr){5-5}\cmidrule(lr){6-6}
  %  \cmidrule(lr){1-10}
  %  WN18  & & & 141,442 & 5,000 & 5,000 \\%& 5,760,450 & 10,000 & 10,000 & 13,133 \\
  %  FB15k & & & 483,142 & 50,000 & 59,071 \\%& 7,861,321 & 30,000 & 30,000 & 59,586 \\
  %  YAGO3-10 & \\
  %  \bottomrule
  %\end{tabular}
  %}
\end{table}

%%%%%%%%%%%%%%%%%%%%%%%%%%%%%%%%%%%%%%%%%%%%%%%%%%%%%%%%%%%%%%%%%%%%%%%% 

%% Local Variables:
%% Mode: LaTeX
%% TeX-master: "../main"
%% TeX-engine: xetex
%% TeX-PDF-mode: t
%% End:

To verify the effectiveness of B-CP over larger datasets,
we also conducted experiments on the Freebase-music
data\footnote{\url{https://datalab.snu.ac.kr/haten2/}}.
To reduce noises, we removed triples from the data
whose relation and entities occur less than 10 times.
The number of the remaining triples were 18,482,832
which consist of 138 relations and 3,025,684 entities.
We split them randomly into three subsets: 18,462,832
training, 10,000 validation, and 10,000 test triples.
%We randomly generated 20,000 unknown triples and
%added one half of them into validation
%and the other into test triples as negative samples.
We randomly generated 20,000 triples that are not in the knowledge graph, and used them as negative samples; half of them were placed in the validation set, and the other half in the test set.
%\memo{what do you mean by ``unknown'' triples? Here is a suggested rewrite, but I'm not sure if this is what is really meant: We randomly generated 20,000 triples that are not in the knowledge graph, and used them as negative samples; half of them were placed in the validation set, and the other half in the test set.}
%\memo{But then, why no negative examples for training?}
Experiments were conducted under the same hyperparameters and 
negative samples setting
we achieved the best results on the FB15k dataset.
We here report the triple classification accuracy.
%\memo{sounds contradictory; if you use classification accuracy as the performance metric, the ``condition'' is not the same as the small datasets.}
Table~\ref{tab:ldata} gives the results.
As it was with the small datasets, 
the performance of CP~$(D=15)$ was again poor.
Meanwhile, B-CP successfully reduced the model size while achieving better
performance than CP~$(D=200)$.
These results show that B-CP is robust to the data size.
%B-CP also achieved better performance than CP on a large-scale dataset.

%%%%%%%%%%%%%%%%%%%%%%%%%%%%%%%%%%%%%%%%%%%%%%%%%%%%%%%%%%%%%%%%%%%%%%%%

%%% Local Variables:
%%% mode: latex
%%% TeX-PDF-mode: t
%%% TeX-engine: xetex
%%% TeX-master: "main"
%%% End:

\section{Conclusion}
\label{sec:summ}
In this paper,
we showed that it is possible to obtain binary vectors of relations and entities
in knowledge graphs
that take $10$--$20$ times less storage/memory than
the original representations with floating point numbers.
Additionally, with the help of bitwise operations,
the time required for score computation was considerably reduced.
Tensor factorization arises in many machine learning applications
such as item recommendation~\cite{rec} and web link analysis~\cite{web}.
Applying our B-CP algorithm to the analysis of
other relational datasets
%\memo{what is this? I wonder if it's the model that is multilinear, not the dataset?} 
is an interesting avenue for future work.

The program codes for the binarized CP decomposition algorithm
proposed here will be provided on the first author's GitHub page
\footnote{\url{https://github.com/KokiKishimoto/cp\_decomposition.git}}.

%%%%%%%%%%%%%%%%%%%%%%%%%%%%%%%%%%%%%%%%%%%%%%%%%%%%%%%%%%%%%%%%%%%%%%%%

%%% Local Variables:
%%% mode: latex
%%% TeX-PDF-mode: t
%%% TeX-engine: xetex
%%% TeX-master: "main"
%%% End:

\bibliographystyle{splncs04}
\bibliography{main}
\end{document}